\documentclass[10pt,twocolumn,letterpaper]{article}

\usepackage[pagenumbers]{cvpr} 

\usepackage{graphicx}
\usepackage{amsmath}
\usepackage{amssymb}
\usepackage{booktabs}

\usepackage{multirow}
\usepackage[flushleft]{threeparttable} 

\usepackage[pagebackref,breaklinks,colorlinks]{hyperref}

\usepackage[capitalize]{cleveref}
\crefname{section}{Sec.}{Secs.}
\Crefname{section}{Section}{Sections}
\Crefname{table}{Table}{Tables}
\crefname{table}{Tab.}{Tabs.}

\def\cvprPaperID{8276} 
\def\confName{CVPR}
\def\confYear{2022}

\newcommand{\norm}[1]{\left\lVert#1\right\rVert}
\newcommand{\abs}[1]{\left\lvert#1\right\rvert}

\begin{document}

\title{GeoNeRF: Generalizing NeRF with Geometry Priors}

\author{Mohammad Mahdi Johari\\
Idiap Research Institute, EPFL\\
{\tt\small mohammad.johari@idiap.ch}
\and
Yann Lepoittevin\\
ams OSRAM\\
{\tt\small yann.lepoittevin@ams.com}
\and
François Fleuret\\
University of Geneva, EPFL\\
{\tt\small francois.fleuret@unige.ch}
}

\twocolumn[{
\maketitle
\begin{center}
    \vspace{-3ex}
    \captionsetup{type=figure}
    \includegraphics[width=1.0\linewidth]{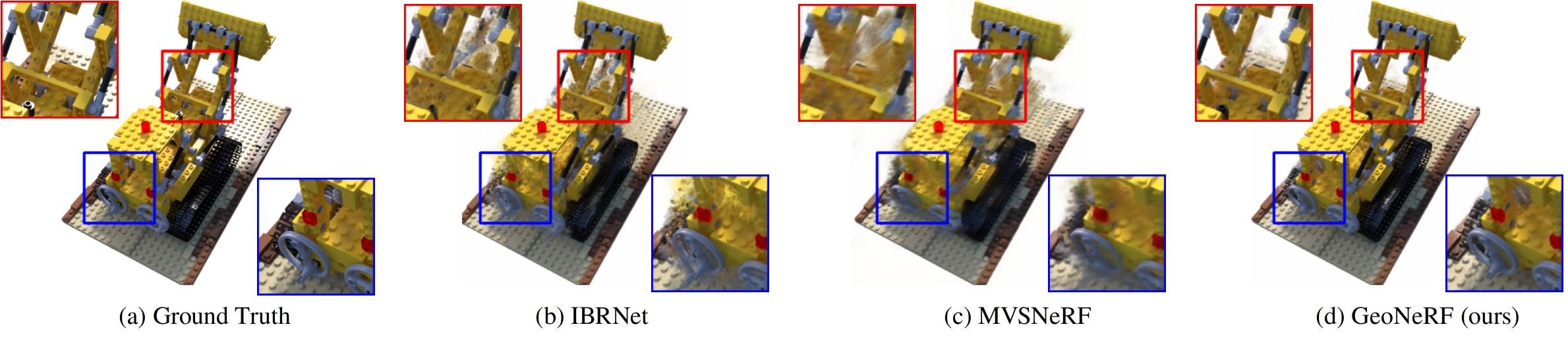}
    \vspace{-4.0ex}
    \captionof{figure}{Our generalizable GeoNeRF model infers complex geometries of objects in a novel scene without per-scene optimization and synthesizes novel images of higher quality than the existing works: IBRNet~\cite{wang2021ibrnet} and MVSNeRF~\cite{chen2021mvsnerf}.}
\end{center}
}]


\begin{abstract}
   We present GeoNeRF, a generalizable photorealistic novel view synthesis method based on neural radiance fields. Our approach consists of two main stages: a geometry reasoner and a renderer. To render a novel view, the geometry reasoner first constructs cascaded cost volumes for each nearby source view. Then, using a Transformer-based attention mechanism and the cascaded cost volumes, the renderer infers geometry and appearance, and renders detailed images via classical volume rendering techniques. This architecture, in particular, allows sophisticated occlusion reasoning, gathering information from consistent source views. Moreover, our method can easily be fine-tuned on a single scene, and renders competitive results with per-scene optimized neural rendering methods with a fraction of computational cost. Experiments show that GeoNeRF outperforms state-of-the-art generalizable neural rendering models on various synthetic and real datasets. Lastly, with a slight modification to the geometry reasoner, we also propose an alternative model that adapts to RGBD images. This model directly exploits the depth information often available thanks to depth sensors. The implementation code is available at \href{https://www.idiap.ch/paper/geonerf}{https://www.idiap.ch/paper/geonerf}.
\end{abstract}

\section{Introduction}
\label{sec:intro}

Novel view synthesis is a long-standing task in computer vision and computer graphics. Neural Radiance Fields (NeRF)~\cite{mildenhall2020nerf} made a significant impact on this research area by implicitly representing the 3D structure of the scene and rendering high-quality novel images. Our work addresses the main drawback of NeRF, which is the requirement to train from scratch for every scene separately. The per-scene optimization of NeRF is lengthy and requires densely captured images from each scene.

Approaches like pixelNeRF~\cite{yu2021pixelnerf}, GRF~\cite{trevithick2021grf}, MINE~\cite{li2021mine}, SRF~\cite{chibane2021stereo}, IBRNet~\cite{wang2021ibrnet}, MVSNeRF~\cite{chen2021mvsnerf}, and recently introduced NeRFormer~\cite{reizenstein2021common} address this issue and generalize NeRF rendering technique to unseen scenes. The common motivation behind such methods is to condition the NeRF renderer with features extracted from source images from a set of nearby views. Despite the generalizability of these models to new scenes, their understanding of the scene geometry and occlusions is limited, resulting in undesired artifacts in the rendered outputs. MVSNeRF~\cite{chen2021mvsnerf} constructs a low-resolution 3D cost volume inspired by MVSNet~\cite{yao2018mvsnet}, which is widely used in the Multi-View Stereo research, to condition and generalize the NeRF renderer. However, it has difficulty rendering detailed images and does not deal with occlusions in a scene. In this work, we take MVSNeRF as a baseline and propose the following improvements.

\begin{itemize}
\item We introduce a geometry reasoner in the form of cascaded cost volumes~(Section~\ref{sec:geometry}) and train it in a semi-supervised fashion~(Section~\ref{sec:loss}) to obtain fine and high-resolution priors for conditioning the renderer.

\item We combine an attention-based model which deals with information coming from different source views at any point in space, by essence permutation invariant, with an auto-encoder network which aggregates information along a ray, leveraging its strong Euclidean and ordering structure~(Section~\ref{sec:renderer}).

\item Thanks to the symmetry and generalizability of our geometry reasoner and renderer, we detect and exclude occluded views for each point in space and use the remaining views for processing that point~(Section~\ref{sec:renderer}). 

\end{itemize}

In addition, with a slight modification to the architecture, we propose an alternate model that takes RGBD images (RGB+Depth) as input and exploits the depth information to improve its perception of the geometry~(Section~\ref{sec:rgbd_method}).

Concurrent to our work, the followings also introduce a generalizable NeRF: RGBD-Net~\cite{nguyen2021rgbd} builds a cost volume for the target view instead of source views, NeuralMVS\cite{rosu2021neuralmvs} proposes a coarse to fine approach to increase speed, and NeuRay~\cite{liu2021neural} proposes a method to deal with occlusions.

\section{Related Work} \label{sec:related}
\noindent\textbf{Multi-View Stereo.} The purpose of Multi-View Stereo (MVS) is to estimate the dense representation of a scene given multiple overlapping images. This field has been extensively studied: first, with now called traditional methods \cite{kolmogorov2002multi, de1999poxels, furukawa2009accurate, schonberger2016pixelwise} and more recently with methods relying on deep learning such as MVSNet~\cite{yao2018mvsnet}, which outperformed the traditional ones. MVSNet estimates the depth from multiple views by extracting features from all images, aggregating them into a variance-based cost volume after warping each view onto the reference one, and finally, post-processing the cost volumes with a 3D-CNN. The memory needed to post-process the cost volume being the main bottleneck of~\cite{yao2018mvsnet}, R-MVSNet~\cite{yao2019recurrent} proposed regularizing the cost volume along the depth direction with gated recurrent units while slightly sacrificing accuracy. To further reduce the memory impact, \cite{gu2020cascade,cheng2020deep,yang2020cost} proposed cascaded architectures, where the cost volume is built at gradually finer scales, with the depth output computed in a coarse to fine manner without any compromise on the accuracy. Replacing the variance-based metric with group-wise correlation similarity is another approach to further decrease the memory usage of MVS networks~\cite{xu2020learning}. We found MVS architectures suitable for inferring the geometry and occlusions in a scene and conditioning a novel image renderer. \\  

\noindent\textbf{Novel View Synthesis.} Early work on synthesizing novel views from a set of reference images was done by blending reference pixels according to specific weights~\cite{debevec1996modeling, levoy1996light}. The weights were computed according to ray-space proximity~\cite{levoy1996light} or approximated geometry~\cite{buehler2001unstructured, debevec1996modeling}. To improve the computed geometry, some used the optical flow~\cite{casas20154d, du2018montage4d} or soft blending~\cite{penner2017soft}. Others synthesized a radiance field directly on a mesh~\cite{debevec1998efficient, huang2020adversarial} or on a point cloud~\cite{aliev2020neural, meshry2019neural}. An advantage of these methods is that they can synthesize new views with a small number of references, but their performance is limited by the quality of 3D reconstruction~\cite{jancosek2011multi, schonberger2016structure}, and problems often arise in low-textured or reflective regions where stereo reconstruction tends to fail. Leveraging CNNs to predict volumetric representations stored in voxel grids~\cite{kalantari2016learning, penner2017soft, henzler2020learning} or Multi-Plane Images~\cite{flynn2016deepstereo, zhou2018stereo, srinivasan2019pushing,flynn2019deepview} produces photo-realistic renderings. Those methods rely on discrete volumetric representations of the scenes limiting their outputs' resolution. They also need to be trained on large datasets to store large numbers of samples resulting in extensive memory overhead. \\

\noindent\textbf{Neural Scene Representations.} Recently, using neural networks to represent the geometry and appearance of scenes has allowed querying color and opacity in continuous space and viewing directions. NeRF~\cite{mildenhall2020nerf} achieves impressive results for novel view synthesis by optimizing a 5D neural radiance field for a scene. Building upon NeRF many improvements were made \cite{park2021nerfies, li2021neural, martin2021nerf, peng2021neural, schwarz2020graf, srinivasan2021nerv, rockwell2021pixelsynth, DeVries_2021_ICCV, Barron_2021_ICCV}, but the network needs to be optimized for hours or days for each new scene. Later works, such as GRF~\cite{trevithick2021grf}, pixelNeRF~\cite{yu2021pixelnerf} and MINE~\cite{li2021mine}, try to synthesize novel views with very sparse inputs, but their generalization ability to challenging scenes with complex specularities is highly restricted.
MVSNeRF~\cite{chen2021mvsnerf} proposes to use a low-resolution plane-swept cost volume to generalize rendering to new scenes with as few as three images without retraining. Once the cost volume is computed, MVSNeRF uses a 3D-CNN to aggregate image features. This 3D-CNN resembles the generic view interpolation function presented in IBRNet~\cite{wang2021ibrnet} that allows rendering novel views on unseen scenes with few images. Inspired by MVSNeRF, our work first constructs cascaded cost volumes per source view and then aggregates the cost volumes of the views in an attention-based approach. The former allows capturing high-resolution details, and the latter addresses occlusions.

\section{Method}
\label{sec:method}
\begin{figure*}
    \begin{center}
        \includegraphics[width=0.99\linewidth]{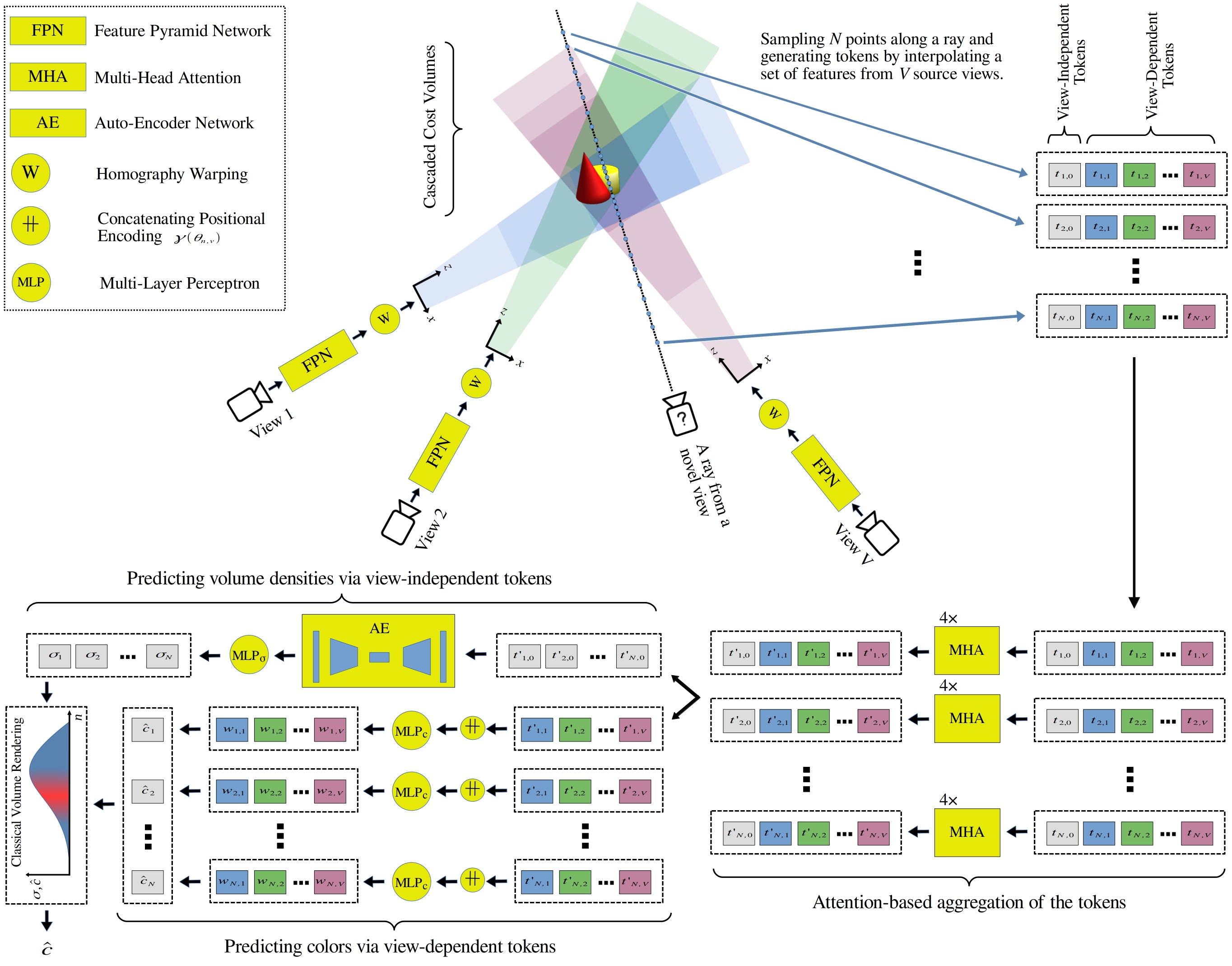}
    \end{center}
    \vspace{-3.25ex}
   \caption{The overview of GeoNeRF. 2D feature pyramids are first generated via Feature Pyramid Network (FPN)~\cite{lin2017feature} for each source view~$v$. We then construct cascaded cost volumes at three levels for each view by homography warping of its nearby views (see Section~\ref{sec:geometry}). Guided by the distribution of the cascaded cost volumes in the 3D space, $N=N_{c}+N_{f}$ points~$\{x_{n}\}_{n=1}^{N}$ are sampled along a ray for a novel pose (see Section~\ref{sec:sampling}). By interpolating both 2D and 3D features~($\boldsymbol{f_{n,v}^{(0)}}$, $\{\boldsymbol{\Phi_{n,v}^{(l)}}\}_{l=0}^{2}$) from FPN and cascaded cost volumes for each sample point~$x_{n}$, one view independent token~$\boldsymbol{t_{n,0}}$ and $V$ view-dependent tokens~$\{\boldsymbol{t_{n,v}}\}_{v=1}^{V}$ are generated. These $V+1$ tokens go through four stacked Multi-Head Attention (MHA) layers and yield more refined tokens~$\{\boldsymbol{t'_{n,v}}\}_{v=0}^{V}$. The MHA layers are shared among all sample points on a ray. Thereafter, the view-independent tokens~$\{\boldsymbol{t'_{n,0}}\}_{n=1}^{N}$ are regularized and aggregated along the ray samples through the AE network, and volume densities~$\{\boldsymbol{\sigma_{n}}\}_{n=1}^{N}$ of the sampled points are estimated. Other tokens~$\{\boldsymbol{t'_{n,v}}\}_{v=1}^{V}$, supplemented with the positional encodings~$\{\gamma(\theta_{n,v})\}_{v=1}^{V}$, predict the color weights~$\{\boldsymbol{w_{n,v}}\}_{v=1}^{V}$ with respect to source views, and the color~$\boldsymbol{\hat{c}_{n}}$ of each point is estimated in a weighted sum fashion (see Section~\ref{sec:renderer}). Finally, the color of the ray~$\boldsymbol{\hat{c}}$ is rendered using classical volume rendering.}
   \vspace{-2.5ex}
    \label{fig:arch}
\end{figure*}

We use volume rendering techniques to synthesize novel views given a set of input source views. Our proposed architecture is presented in Figure~\ref{fig:arch}, and the following sections provide details of our method.

\subsection{Geometry Reasoner} \label{sec:geometry}
Given a set of $V$ nearby views $\{I_{v}\}_{v=1}^{V}$ with size $H \times W$, our geometry reasoner constructs cascaded cost volumes for each input view individually, following the same approach in CasMVSNet~\cite{gu2020cascade}. First, each image goes through a Feature Pyramid Network (FPN)~\cite{lin2017feature} to generate semantic 2D features at three different scale levels.
\begin{equation}
    \boldsymbol{f^{(l)}_{v}}=\text{FPN} \left( I_v \right) \in \mathbb{R}^{\frac{H}{2^l} \times \frac{W}{2^l} \times 2^lC} \phantom{0} \forall l \in \{0, 1, 2\}
\end{equation}
where FPN is the Feature Pyramid Network, $C$ is the channel dimension at level 0, and $l$ indicates the scale level. Once 2D features are generated, we follow the same approach in CasMVSNet to construct plane sweeps and cascaded cost volumes at three levels via differentiable homography warping. CasMVSNet originally estimates depth maps~$\boldsymbol{\hat{D}^{(l)}}$ of the input images at three levels. The coarsest level ($l=2$) consists of $D^{(2)}$ plane sweeps covering the whole depth range in the camera's frustum. Then, subsequent levels narrow the hypothesis range (decrease $D^{(l)}$) but increase the spatial resolution of each voxel by creating $D^{(l)}$ finer plane sweeps on both sides of the estimated depths from the previous level. As a result, the finer the cost volume is, the thinner the depth range it covers. We make two modifications to the CasMVSNet architecture and use it as the geometry reasoner. Firstly, we provide an additional output head to the network to produce multi-level semantic 3D features~$\boldsymbol{\Phi^{(l)}}$ along with the estimated depth maps~$\boldsymbol{\hat{D}^{(l)}}$. Secondly, we replace the variance-based metric in CasMVSNet with \textbf{group-wise correlation similarity} from~\cite{xu2020learning} to construct light-weight volumes. Group-wise correlation decreases memory usage and  inference time.

To be more specific, for each source view~$I_{v}$, we first form a set of its nearby views~$\Gamma_{v}$. Then, by constructing $D^{(l)}$ depth plane sweeps and homography warping techniques, we create multi-level cost volumes~$\boldsymbol{P^{(l)}_{v}}$ from 2D feature pyramids~$\boldsymbol{f^{(l)}}$ of images in $\Gamma_{v}$ using the group-wise correlation similarity metric from~\cite{xu2020learning}. We finally further process and regularize the cost volumes using 3D hourglass networks $R^{(l)}_{3D}$ and generate depth maps~$\boldsymbol{\hat{D}^{(l)}_{v}} \in \mathbb{R}^{\frac{H}{2^l} \times \frac{W}{2^l} \times 1}$ and 3D feature maps~$\boldsymbol{\Phi^{(l)}_{v}} \in \mathbb{R}^{D^{(l)} \times \frac{H}{2^l} \times \frac{W}{2^l} \times C}$:
\begin{equation}
    \boldsymbol{\hat{D}^{(l)}_{v}}, \boldsymbol{\Phi^{(l)}_{v}}=R^{(l)}_{3D} \left( \boldsymbol{P^{(l)}_{v}} \right)  \phantom{0} \forall l \in \{0, 1, 2\}
\end{equation}

\subsection{Sampling Points on a Novel Ray} \label{sec:sampling}

Once the features from the geometry reasoner are generated, we render novel views with the ray casting approach. For each camera ray at a novel camera pose, we first sample $N_{c}$ points along the ray uniformly to cover the whole depth range. Furthermore, we estimate a stepwise probability density function $p_{0}(x)$ along the ray representing the probability that a point $x$ is covered by a full-resolution partial cost volume $\boldsymbol{P^{(0)}}$. Voxels inside the thinnest, full-resolution cost volumes $\{\boldsymbol{P_{v}^{(0)}}\}_{v=1}^{V}$ contain the most valuable information about the surfaces and geometry. Therefore, we sample $N_{f}$ more points from $p_{0}(x)$ distribution. Unlike previous works\cite{mildenhall2020nerf,yu2021pixelnerf, wang2021ibrnet, reizenstein2021common, arandjelovic2021nerf} that require training two networks simultaneously (one coarse and one fine network) and rendering volume densities from the coarse network to resample more points for the fine one, we sample a mixture of $N=N_{c} + N_{f}$ valuable points before rendering the ray without any computation overhead or network duplication thanks to the design of our geometry reasoner.

\subsection{Renderer} \label{sec:renderer}
For all sample points $\{x_{n}\}_{n=1}^{N}$, we interpolate the full-resolution 2D features $\boldsymbol{f_{n,v}^{(0)}}$ and the three-level 3D features $\{\boldsymbol{\Phi_{n,v}^{(l)}}\}_{l=0}^{2}$ from all source views. We also define an occlusion mask $M_{n,v}$ for each point $x_{n}$ with respect to each view $v$. Formally, if a point $x_{n}$ stands behind the estimated full-resolution depth map $\boldsymbol{\hat{D}^{(0)}_{v}}$ (being occluded) or the projection of $x_{n}$ to the camera plane of view $v$ lies outside of the image plane (being outside of the camera frustum), we set $M_{n,v} = 0$ and discard view $v$ from the rendering process of point $x_{n}$. Next, we create a view-independent token $t_{n,0}$ and $V$ view-dependent tokens $\{t_{n,v}\}_{v=1}^{V}$ by utilizing the interpolated features for each point $x_{n}$:
\begin{equation}\begin{aligned}
\boldsymbol{t_{n,v}} &= \text{LT} \left( \left[\boldsymbol{f^{(0)}_{n,v}}; \{ \boldsymbol{\Phi^{(l)}_{n,v}} \}_{l=0}^{2} \right] \right) \phantom{0} \forall v \in \{1, ..., V\}
\\
\boldsymbol{t_{n,0}} &= \text{LT} \left( \left[mean\{\boldsymbol{f^{(0)}_{n,v}}\}_{v=1}^{V}; var\{\boldsymbol{f^{(0)}_{n,v}}\}_{v=1}^{V}\right] \right)
\end{aligned}\end{equation}
where $\text{LT}(\cdot)$ and $[\cdot \phantom{1}; \phantom{1} \cdot]$ denote respectively linear transformation and concatenation. $\boldsymbol{t_{n,0}}$ could be considered as a global understanding of the scene at point $x_{n}$, while $\boldsymbol{t_{n,v}}$ represents the understanding of the scene from source view $v$. The global and view-dependent tokens are aggregated through four stacked Multi-Head Attention (MHA) layers, which are introduced in Transformers~\cite{dosovitskiy2020image, vaswani2017attention}:
\begin{equation}
    \{\boldsymbol{t'_{n,v}}\}_{v=0}^{V} = \phantom{}^{4 \times} \text{MHA} \left(\boldsymbol{t_{n,0}}, \{\boldsymbol{t_{n,v}}, M_{n,v}\}_{v=1}^{V} \right)
\end{equation}

Our MHA layers also take the occlusion masks $M_{n,v}$ as inputs and force the occluded views' attention scores to zero to prevent them from contributing to the aggregation.

The global view-independent output tokens $\{\boldsymbol{t'_{n,0}}\}_{n=1}^{N}$ now have access to all necessary data to learn the geometry of the scene and estimate volume densities. We further regularize these tokens through an auto-encoder-style (AE) network in the ray dimension ($n$). The AE network learns the global geometry along the ray via convolutional layers and predicts more coherent volume densities~$\boldsymbol{\sigma_{n}}$:
\begin{equation}
    \{\boldsymbol{\sigma_{n}}\}_{n=1}^{N} = \text{MLP}_{\sigma} \left( \text{AE} \left(\{\boldsymbol{t'_{n,0}}\}_{n=1}^{N} \right) \right)
\end{equation}
where $\text{MLP}_{\sigma}$ is a simple two-layer perceptron. We argue that convolutionally processing the tokens with the AE network along the ray dimension ($n$) is a proper inductive bias and significantly reduces the computation resources compared to methods like IBRNet~\cite{wang2021ibrnet} and NeRFormer~\cite{reizenstein2021common}, which employ an attention-based architecture because the geometry of a scene is naturally continuous, and accordingly, closer points are more likely related.

View-dependent tokens~\{$\boldsymbol{t'_{n,v}}\}_{v=1}^{V}$, together with two additional inputs, are used for color prediction. We project each point $x_{n}$ to source views' image planes and interpolate the color samples $c_{n,v}$. We also calculate the angle between the novel camera ray and the line that passes through the camera center of source view $v$ and $x_{n}$. This angle $\theta_{n,v}$ represents the similarity between the camera pose of source view $v$ and the novel view. Each point's color is estimated via a weighted sum of the non-occluded views' colors: 
\begin{align}
\boldsymbol{w_{n,v}} &= \text{Softmax} \left ( \big \{\text{MLP}_c \left ([\boldsymbol{t'_{n,v}}; \gamma(\theta_{n,v})] \right ),M_{n,v} \big \}_{v=1}^{V} \right ) \notag
\\
\boldsymbol{\hat{c}_{n}} &= \sum_{v=1}^{V} \boldsymbol{w_{n,v}}c_{n,v} \phantom{0} \forall n \in \{1, 2, ..., N\}
\end{align}
where $\gamma(\cdot)$ is the sinusoidal positional encoding proposed in~\cite{mildenhall2020nerf}, and $\text{MLP}_{c}$ is a simple two-layer perceptron. The Softmax function also takes the occlusion masks $M_{n,v}$ as input to exclude occluded views.

Once volume densities and colors are predicted, our model renders, as in NeRF~\cite{mildenhall2020nerf}, the color of the camera ray at a novel pose using the volume rendering approach:
\begin{equation}
\boldsymbol{\hat{c}} = \sum_{n=1}^{N} \exp \left( -\sum_{k=1}^{n-1} \boldsymbol{\sigma_{k}} \right) \left( 1 - \exp \left(-\boldsymbol{\sigma_{n}} \right)\right)\boldsymbol{\hat{c}_{n}}
\end{equation}
In addition to the rendered color, our model also outputs the estimated depth $\boldsymbol{\hat{d}}$ for each ray:
\begin{equation}
    \boldsymbol{\hat{d}} = \sum_{n=1}^{N} \exp \left(-\sum_{k=1}^{n-1} \boldsymbol{\sigma_{k}} \right) \left(1 - \exp \left( -\boldsymbol{\sigma_{n}} \right) \right)z_{n}
\end{equation}
where $z_{n}$ is the depth of point~$x_{n}$ with respect to the novel pose. This auxiliary output is helpful for training and supervising our generalizable model (see Section~\ref{sec:loss}).

\subsection{Loss Functions} \label{sec:loss}
The primary loss function when we train our generalizable model on various scenes and the \textbf{only} loss when we fine-tune it on a specific scene is the mean squared error between the rendered colors and ground truth pixel colors:
\begin{equation}
    \mathcal{L}_{c} = \frac{1}{|R|} \sum_{r \in R} \norm{ \boldsymbol{\hat{c}}(r) - c_{gt}(r) }^2
\end{equation}
where $R$ is the set of rays in each training batch and $c_{gt}$ is the ground truth color.

DS-NeRF~\cite{deng2021depth} shows that depth supervision can help NeRF train faster with fewer input views. Moreover, numerous works~\cite{kaya2022neural, Oechsle2021ICCV, wang2021neus, yariv2021volume} show that despite the high-quality color rendering, NeRF has difficulty reconstructing 3D geometry and surface normals. Accordingly, for training samples coming from datasets with ground truth depths, we also output the predicted depth~$\boldsymbol{\hat{d}}$ for each ray and supervise it if the ground truth depth of that pixel is available:
\begin{equation}
    \mathcal{L}_{d} = \frac{1}{|R_{d}|} \sum_{r \in R_{d}}  \norm{ \boldsymbol{\hat{d}}(r) - d_{gt}(r)}_{s1}
\end{equation}
where $R_{d}$ is the set of rays from samples with ground truth depths and $d_{gt}$ is the pixel ground truth depth and $||\cdot||_{s1}$ is the smooth $L_1$ loss.

Lastly, we supervise cascaded depth estimation networks in our geometry reasoner. For datasets with ground truth depth, the loss is defined as:
\begin{equation}
    \mathcal{L}^{(l)}_{D} = \frac{2^{- l}}{|V|} \sum_{v=1}^{V} \left < \norm{ \boldsymbol{\hat{D}^{(l)}_{v}} - D^{(l)}_{v}}_{s1} \right >
\end{equation}
where $D^{(l)}_{v}$ is the ground truth depth map of view $v$ resized to scale level $l$, and $< \cdot >$ denotes averaging over all pixels. For training samples without ground truth depths, we self-supervise the depth maps. We take the rendered ray depths as pseudo-ground truth and warp their corresponding colors and estimated depths from all source views using camera transformation matrices. If the ground truth pixel color of a ray is consistent with the warped color of a source view, and it is located in a textured neighborhood, we allow $\boldsymbol{\hat{d}}$ to supervise the geometry reasoner for that view. Formally:
\begin{align}
        & \mathcal{L}^{(l)}_{D} = \frac{2^{- l}}{|V||R|} \sum_{v=1}^{V} \sum_{r \in R} M_{v}(r) \norm{ \boldsymbol{\hat{D}^{(l)}_{v}}(r_v) - \boldsymbol{\hat{d}}(r_v)}_{s1} \notag
        \\
        & \text{where } \phantom{0} r_{v} = T_{\rightarrow v}\left(r, \boldsymbol{\hat{d}}(r)\right)
        \\
        & \text{and } \phantom{0} M_{v}(r) = 
        \begin{cases} 
            1 & \text{ if } \phantom{0}
            \parbox[t]{0.6\linewidth}{
                $\abs{ I_{v}(r_v) - c_{gt}(r)} < \epsilon_c $\\
                and $V_5 \left( I_{v}(r_v) \right) > \epsilon_t$
            }
            \\
            0 & \text{ otherwise }
        \end{cases} \notag
\end{align}
Given a ray $r$ at a novel pose with rendered depth $\boldsymbol{\hat{d}}(r)$, $T_{\rightarrow v}\left(r, \boldsymbol{\hat{d}}(r)\right)$ transforms the ray to its correspondent ray from source view $v$ using camera matrices. $\boldsymbol{\hat{d}}(r_v)$ denotes the rendered depth of the correspondent ray with respect to source view $v$, and $M_{v}(r)$ validates the texturedness and color consistency. We keep pixels whose variance~$V_5(\cdot)$ in their $5 \times 5$ pixels neighborhood is higher than $\epsilon_t$, and whose color differs less than $\epsilon_c$ from the color of the ray~$r$. The aggregated loss function for our generalizable model is:
\begin{equation}
    \mathcal{L} = \mathcal{L}_{c} + 0.1 \mathcal{L}_{d} + \lambda \sum_{l=0}^{2}\mathcal{L}^{(l)}_{D}
\end{equation}
where $\lambda$ is 1.0 if the supervision is with ground truth depths and is 0.1 if it is with pseudo-ground truth rendered depths. For \textbf{fine-tuning} on a single scene, regardless of the availability of depth data, we only use $\mathcal{L}_{c}$ as the loss function.

\subsection{Compatibility with RGBD data} \label{sec:rgbd_method}
Concerning the ubiquitousness of the embedded depth sensors in devices nowadays, we also propose an RGBD compatible model, $\text{GeoNeRF}_{\text{+D}}$, by making a small modification to the geometry reasoner. We assume an \textbf{incomplete}, low-resolution, noisy depth map~$D_{v} \in \mathbb{R}^{\frac{H}{4} \times \frac{W}{4} \times 1}$ is available for each source view $v$. When we construct the coarsest cost volume~$\boldsymbol{P^{(2)}_{v}}$ with $D^{(2)}$ depth planes, we also construct a binary volume~$B_{v} \in \mathbb{R}^{D^{(2)} \times \frac{H}{4} \times \frac{W}{4} \times 1}$ and concatenate it with $\boldsymbol{P^{(2)}_{v}}$ before feeding them to the $R^{(2)}_{3D}$ network:
\begin{equation}
    B_{v}(d,h,w) = \begin{cases} 
    1 &\mbox{if } Q \left( D_{v} \left(h,w \right) \right) \equiv d \\
    0 & \mbox{otherwise} 
    \end{cases}
\end{equation}
where $Q(\cdot)$ maps and quantizes real depth values to depth plane indices. $B_{v}$ plays the role of coarse guidance of the geometry in $\text{GeoNeRF}_{\text{+D}}$. As a result of this design, the model is robust to quality and sparsity of the depth inputs.

\begin{table*}[!t]
    \begin{center}
        \begin{threeparttable}
            \begin{tabular}{l|c|ccc|ccc|ccc}
            \hline
            \multirow{2}{*}{Method} & \multirow{2}{*}{Settings} & \multicolumn{3}{c|}{DTU MVS\tnote{\dag} ~\cite{jensen2014large}} &  \multicolumn{3}{c|}{Realistic Synthetic~\cite{mildenhall2020nerf}} & \multicolumn{3}{c}{Real Forward Facing~\cite{mildenhall2019llff}}\\
            &  & \small PSNR$\uparrow$ & \small SSIM$\uparrow$ & \small LPIPS$\downarrow$ & \small PSNR$\uparrow$ & \small SSIM$\uparrow$ & \small LPIPS$\downarrow$ & \small PSNR$\uparrow$ & \small SSIM$\uparrow$ & \small LPIPS$\downarrow$ \\
            \hline
            pixelNeRF~\cite{yu2021pixelnerf} &  & 19.31 & 0.789 & 0.382 & \phantom{0}7.39 & 0.658 & 0.411 & 11.24 & 0.486 & 0.671 \\
            IBRNet~\cite{wang2021ibrnet} & No per-scene & 26.04 & 0.917 & 0.190 & \underline{25.49} & \underline{0.916} & \underline{0.100} & \underline{25.13} & \underline{0.817} & \underline{0.205} \\
            MVSNeRF~\cite{chen2021mvsnerf} & Optimization & \underline{26.63} & \underline{0.931} & \underline{0.168} & 23.62 & 0.897 & 0.176 & 21.93 & 0.795 & 0.252 \\
            GeoNeRF &  & \textbf{31.34} & \textbf{0.959} & \textbf{0.060} & \textbf{28.33} & \textbf{0.938} & \textbf{0.087} & \textbf{25.44} & \textbf{0.839} & \textbf{0.180} \\
            \hline
            IBRNet~\cite{wang2021ibrnet} &  & 31.35 & 0.956 & \underline{0.131} & 28.14 & 0.942 & 0.072 & \textbf{26.73} & 0.851 & 0.175 \\
            MVSNeRF\tnote{\ddag}~\cite{chen2021mvsnerf} & \multirow{2}{*}{Per-scene} & 28.50 & 0.933 & 0.179 & 27.07 & 0.931 & 0.168 & 25.45 & \textbf{0.877} & 0.192 \\
            NeRF~\cite{mildenhall2020nerf} & \multirow{2}{*}{Optimization} & 27.01 & 0.902 & 0.263 & \textbf{31.01} & 0.947 & 0.081 & 26.50 & 0.811 & 0.250 \\
            $\text{GeoNeRF}_{\text{10k}}$ &  & \textbf{31.66} & \textbf{0.961} & \textbf{0.059} & \underline{30.42} & \textbf{0.956} & \textbf{0.055} & \underline{26.58} & \underline{0.856} & \textbf{0.162} \\
            $\text{GeoNeRF}_{\text{1k}}$ &  & \underline{31.52} & \underline{0.960} & \textbf{0.059} & 29.83 & \underline{0.952} & \underline{0.061} & 26.31 & 0.852 & \underline{0.164} \\
            \hline
            \end{tabular}
            \begin{tablenotes}
                \small
                \item[\dag] The evaluations of the baselines on the DTU MVS dataset~\cite{jensen2014large} are borrowed from the paper MVSNeRF~\cite{chen2021mvsnerf}. Also, metrics are calculated for all methods on this dataset on foreground pixels, whose ground truth depths stand inside the scene bound.
                \item[\ddag] For fine-tuning, MVSNeRF~\cite{chen2021mvsnerf} discards CNNs and directly optimizes 3D features. Direct optimization without regularization severely suffers from overfitting which is not reflected in their reported accuracy because their evaluation is done on a custom test set instead of the standard one. \textit{E.g.}, their PSNR on NeRF dataset~\cite{mildenhall2020nerf} would drop from \textbf{27.07} to \textbf{20.02} if it was evaluated on the standard test set.
            \end{tablenotes}
        \end{threeparttable}
    \end{center}
    \vspace{-2.5ex}
    \caption{Quantitative comparison of our proposed GeoNeRF with existing generalizable NeRF models in terms of PSNR (higher is better), SSIM~\cite{wang2004image} (higher is better), and LPIPS~\cite{zhang2018unreasonable} (lower is better) metrics. Highlights are \textbf{best} and \underline{second best}. GeoNeRF is superior to the existing approaches in all experiments in which the methods are evaluated without any per-scene optimization (the top row). Notably, GeoNeRF outperforms others with a significant margin on datasets with relatively sparser source views: DTU MVS~\cite{jensen2014large} and NeRF realistic synthetic~\cite{mildenhall2020nerf}. In particular, our method generalizes outstandingly well on the NeRF synthetic dataset~\cite{mildenhall2020nerf}, although our training dataset only contains real scenes which greatly differ from the synthetic scenes. The bottom row of the table presents the evaluation of the methods when they are fine-tuned on each scene separately, as well as a comparison with vanilla NeRF~\cite{mildenhall2020nerf}, which is per-scene optimized for 200k$\textrm{--}$500k iterations. After fine-tuning for only 10k iterations, our $\text{GeoNeRF}_{\text{10k}}$ produces competitive results with NeRF. Remarkably, even after fine-tuning for 1k iterations (approximately one hour on a single V100 GPU), $\text{GeoNeRF}_{\text{1k}}$ reaches 98.15\% of the $\text{GeoNeRF}_{\text{10k}}$'s performance on average, which is another evidence for efficient convergence of our model on novel scenes.}
    \label{table:quantitative}
    \vspace{-2.0ex}
\end{table*}

\section{Experiments}
\label{sec:experiments}
\noindent\textbf{Training datasets.} We train our model on the real DTU dataset~\cite{jensen2014large} and real forward-facing datasets from LLFF~\cite{mildenhall2019llff} and IBRNet~\cite{wang2021ibrnet}. We exclude views with incorrect exposure from the DTU dataset~\cite{jensen2014large} following the practice in pixelNeRF~\cite{yu2021pixelnerf} and use the same 88 scenes for training as in pixelNeRF~\cite{yu2021pixelnerf} and MVSNeRF~\cite{chen2021mvsnerf}. Ground truth depths of DTU~\cite{jensen2014large}, provided by~\cite{yao2018mvsnet}, are the only data that is directly used for depth supervision. For samples from forward-facing datasets (35 scenes from LLFF~\cite{mildenhall2019llff} and 67 scenes from IBRNet~\cite{wang2021ibrnet}), depth supervision is in the self-supervised form. \\

\noindent\textbf{Evaluation datasets.} We evaluate our model on the 16 test scenes of DTU MVS~\cite{jensen2014large}, the 8 test scenes of real forward-facing dataset from LLFF~\cite{mildenhall2019llff}, and the 8 scenes in NeRF realistic synthetic dataset~\cite{mildenhall2020nerf}. We follow the same evaluation protocol in NeRF~\cite{mildenhall2020nerf} for the synthetic dataset~\cite{mildenhall2020nerf} and LLFF dataset~\cite{mildenhall2019llff} and the same protocol in MVSNeRF~\cite{chen2021mvsnerf} for the DTU dataset~\cite{jensen2014large}. Specifically, for LLFF~\cite{mildenhall2019llff}, we hold out $\frac{1}{8}$ of the views of the unseen scenes, and for DTU~\cite{jensen2014large}, we hold out 4 views of the unseen scenes for testing and leave the rest for fine-tuning.\\

\noindent\textbf{Implementation details.} We train the generalizable GeoNeRF for 250k iterations. For each iteration, one scene is randomly sampled, and 512 rays are randomly selected as the training batch. For training on a specific scene, we only fine-tune the model for 10k iterations ($\text{GeoNeRF}_{\text{10k}}$), in contrast to NeRF~\cite{mildenhall2020nerf} that requires 200k$\textrm{--}$500k optimization steps per scene. Since our renderer's architecture is agnostic to the number of source views, we flexibly employ a different number of source views~$V$ for training and evaluation to reduce memory usage. We use $V=6$ source views for training the generalizable model and $V=9$ for evaluation. For fine-tuning, we select $V$ based on the images' resolution and available GPU memory. Specifically, we set $V=9$ for the DTU dataset~\cite{jensen2014large} and $V=7$ for the other two datasets. We fix the number of sample points on a ray to $N_{c} = 96$ and $N_{f} = 32$ for all scenes. We utilize Adam~\cite{adam} as the optimizer with a learning rate of $5 \times 10^{-4}$ for training the generalizable model and a learning rate of $2 \times 10^{-4}$ for fine-tuning. A cosine learning rate scheduler~\cite{sgdr} without restart is also applied to the optimizer. For the details of our networks' architectures, refer to the supplementary.

\begin{figure*}
    \begin{center}
        \includegraphics[width=1\linewidth]{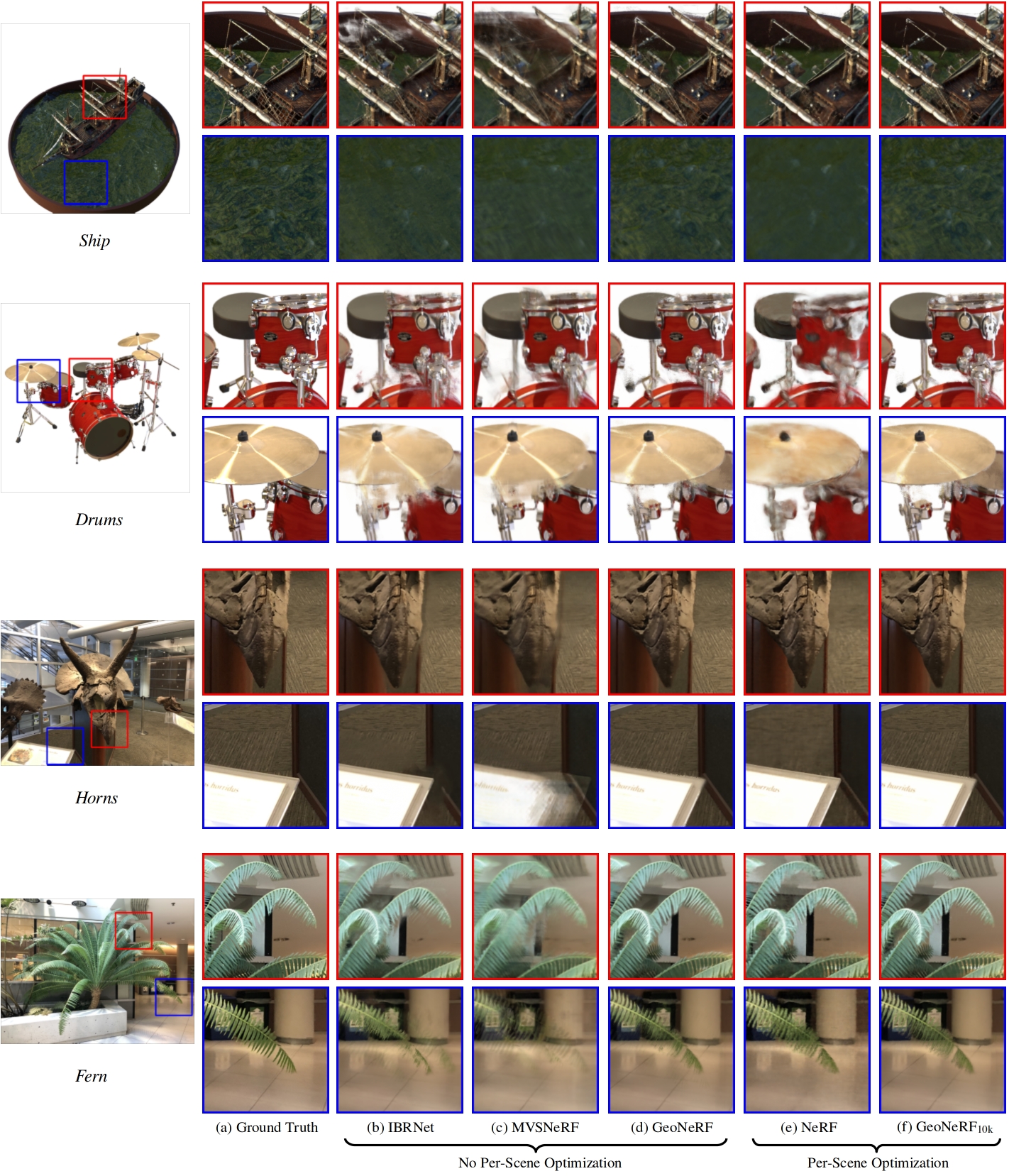}
    \end{center}
    \vspace{-3ex}
   \caption{Qualitative comparison of the methods on the NeRF synthetic dataset~\cite{martin2021nerf} (\textit{Ship} and \textit{Drums}) and the real forward-facing dataset~\cite{mildenhall2019llff} (\textit{Horns} and \textit{Fern}). Our proposed GeoNeRF more accurately preserves the details of the scenes while it generates fewer artifacts than IBRNet~\cite{wang2021ibrnet} and MVSNeRF~\cite{chen2021mvsnerf} (\eg the leaves in \textit{Fern} or the cymbal in \textit{Drums}). After fine-tuning our model only for 10k iterations on each individual scene ($\text{GeoNeRF}_{\text{10k}}$), the results are competitive with per-scene optimized vanilla NeRF~\cite{mildenhall2020nerf}. Compared with NeRF, GeoNeRF models produce smoother surfaces in \textit{Drums} and higher quality textures for the water in \textit{Ship} and for the floor in \textit{Horns}.}
    \label{fig:qualitative}
\end{figure*}

\begin{figure}
    \begin{center}
        \includegraphics[width=1\linewidth]{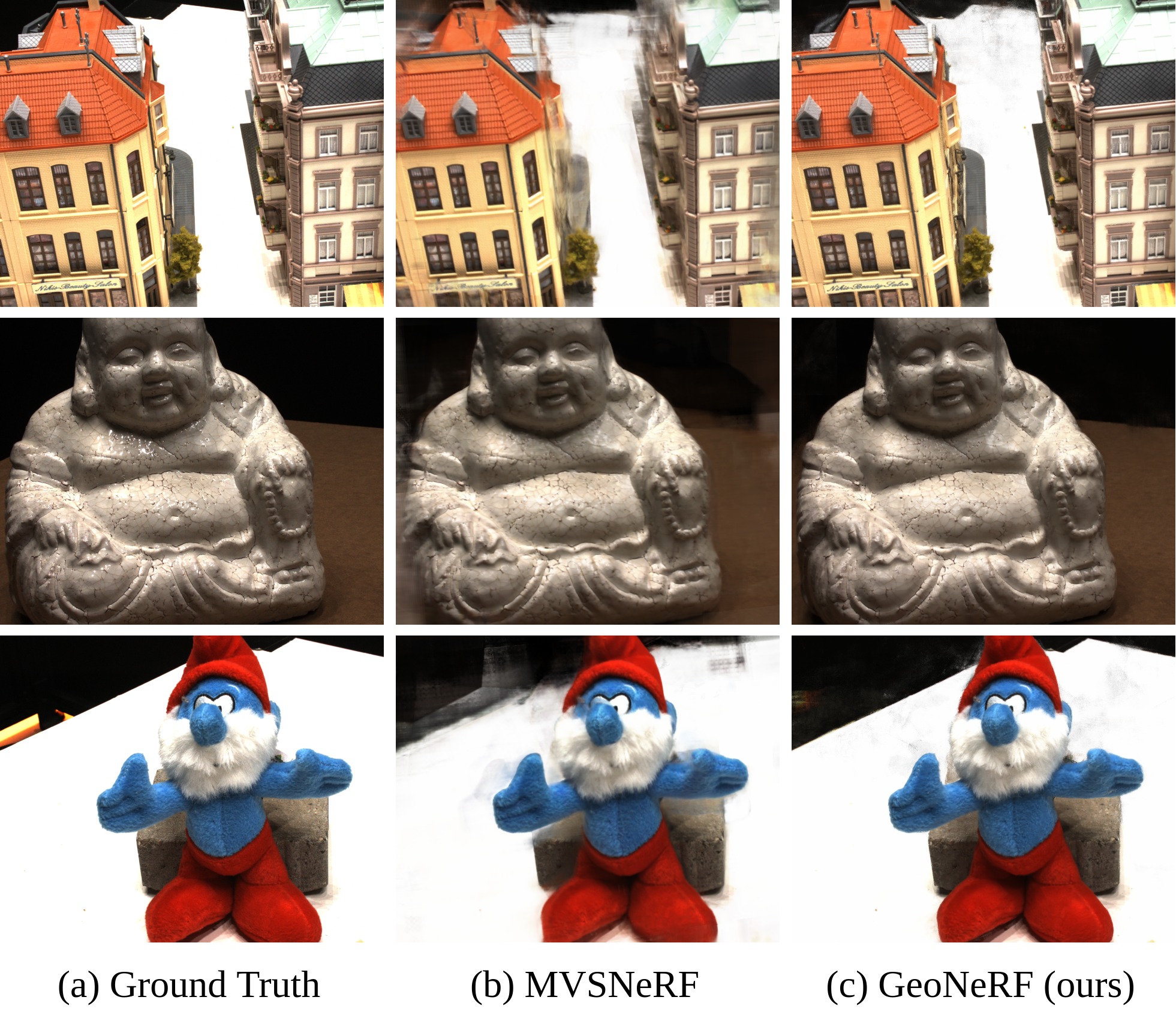}
    \end{center}
    \vspace{-3ex}
   \caption{Qualitative comparison of our generalizable GeoNeRF model with MVSNeRF~\cite{chen2021mvsnerf}, the state-of-the-art model on the DTU dataset~\cite{jensen2014large}. Images are from DTU test scenes. Our method renders sharper images with fewer artifacts.}
    \label{fig:qualitative_dtu}
    \vspace{-2ex}
\end{figure}

\subsection{Experimental Results} We evaluate our model and provide a comparison with the original vanilla NeRF~\cite{mildenhall2020nerf} and the existing generalizable NeRF models: pixelNeRF~\cite{yu2021pixelnerf}, IBRNet~\cite{wang2021ibrnet}, and MVSNeRF~\cite{chen2021mvsnerf}. The authors of NeRFormer~\cite{reizenstein2021common} did not publish their code, did not benchmark their method on NeRF benchmarking datasets, nor test state-of-the-art generalizable NeRF models on their own dataset. They perform on par with NeRF in their experiments with scene-specific optimization and train and test their generalizable model on specific object categories separately. A quantitative comparison is provided in Table~\ref{table:quantitative} in terms of PSNR, SSIM~\cite{wang2004image}, and LPIPS~\cite{zhang2018unreasonable}. The results show the superiority of our GeoNeRF model with respect to the previous generalizable models. Moreover, when fine-tuned on the scenes for only 10k iterations, $\text{GeoNeRF}_{\text{10k}}$ produces competitive results with NeRF, which requires lengthy per-scene optimization. We further show that even after 1k iterations (approximately one hour on a single V100 GPU), $\text{GeoNeRF}_{\text{1k}}$'s results are comparable with NeRF's.

The qualitative comparisons of our model with existing methods on different datasets are provided in Figures~\ref{fig:qualitative} and~\ref{fig:qualitative_dtu}. The images produced by our GeoNeRF model better preserve the details of the scene and contain fewer artifacts. For further qualitative analysis, an extensive ablation study, and limitations of our model, refer to the supplementary.

\subsection{Sensitivity to Source Views} \label{sec:sensitivity}
We conducted two experiments to investigate the robustness of our model to the number and quality of input source views. We first evaluated the impact of the number of source views on our model in Table~\ref{table:n_views}. The results demonstrate the robustness to the sparsity of source views and suggest that GeoNeRF produces high-quality images even with a lower number of source images. Furthermore, we show that our method can operate with both close and distant source views. Table~\ref{table:discard_views} shows the performance when we discard $K$ nearest views to the novel pose and use the remaining source views for rendering. While distant source views are naturally less informative and degrade the quality, our model does not incur a significant decrease in performance.

\begin{table}[!t]
\begin{center}
        \begin{tabular}{l|ccc}
        \hline
        Number of source views & \small PSNR$\uparrow$ & \small SSIM$\uparrow$ & \small LPIPS$\downarrow$  \\
        \hline
        3 & 24.33 & 0.794 & 0.212  \\
        4 & 25.05 & 0.823 & 0.183  \\
        5 & 25.25 & 0.832 & 0.178  \\
        6  & 25.37 & 0.837 & \textbf{0.176} \\
        9 & \textbf{25.44} & \textbf{0.839} & 0.180  \\
        \hline
        IBRNet~\cite{wang2021ibrnet} (10 views) & 25.13 & 0.817 & 0.205  \\
        \hline
        \end{tabular}
    \end{center}
    \vspace{-3.25ex}
    \caption{Quantitative analysis of the robustness of our GeoNeRF to the number of input source views on the LLFF~\cite{mildenhall2019llff} test scenes, besides a comparison with IBRNet~\cite{wang2021ibrnet}, which uses 10 source views.}
    \label{table:n_views}
    \vspace{-0.75ex}
\end{table}

\begin{table}[!t]
	\begin{center}
		\begin{tabular}{lccccc}
			\hline
			$K$: & 0 & 2 & 4 & 6 & 8\\
			\hline
			PSNR$\uparrow$ & \textbf{25.44} & 24.18 & 23.35 & 22.74 & 22.06 \\
			SSIM$\uparrow$ & \textbf{0.839} & 0.813 & 0.791 & 0.770 & 0.747 \\
			LPIPS$\downarrow$ & \textbf{0.180} & 0.212 & 0.235 & 0.253 & 0.276 \\
			\hline
		\end{tabular}
	\end{center}
	
	\vspace*{-3.25ex}
	
	\caption{Quantitative analysis of the sensitivity of our GeoNeRF to discarded first $K$ nearest neighbors on the LLFF~\cite{mildenhall2019llff} test scenes.}
	\label{table:discard_views}
    \vspace{-0.75ex}
\end{table}

\begin{table}[!t]
\begin{center}
        \begin{tabular}{l|ccc}
        \hline
        Model & \small PSNR$\uparrow$ & \small SSIM$\uparrow$ & \small LPIPS$\downarrow$  \\
        \hline
        GeoNeRF & 31.34 & 0.959 & 0.060  \\
        $\text{GeoNeRF}_{\text{+D}}$  & \textbf{31.58} & \textbf{0.961} & \textbf{0.057} \\
        \hline
        \end{tabular}
    \end{center}
    \vspace{-3.25ex}
    \caption{A comparison of the performance of our RGBD compatible $\text{GeoNeRF}_{\text{+D}}$ and original GeoNeRF on DTU~\cite{jensen2014large} test scenes. For the details of this experiment, see Section~\ref{sec:rgbd}.}
    \label{table:rgbd}
    \vspace{-3.0ex}
\end{table}

\subsection{Results with RGBD Images} \label{sec:rgbd}
To evaluate our RGBD compatible model, $\text{GeoNeRF}_{\text{+D}}$, we use the DTU dataset~\cite{jensen2014large} to mimic the real scenario where incomplete, low-resolution depth images accompany RGB images. We feed our model the DTU images with a resolution of $600 \times 800$, while we resize their incomplete depths to $150 \times 200$. The comparison of the performance of GeoNeRF, with and without depth inputs is presented in Table~\ref{table:rgbd}. The results confirm that our $\text{GeoNeRF}_{\text{+D}}$ model adapts to RGBD images and renders higher quality outputs.

\section{Conclusion}
We proposed GeoNeRF, a generalizable learning-based novel view synthesis method that renders state-of-the-art quality images for complex scenes without per-scene optimization. Our method leverages the recent architectures in the multi-view stereo field to understand the scene's geometry and occlusions by constructing cascaded cost volumes for source views. The data coming from the source views are then aggregated through an attention-based network, and images for novel poses are synthesized conditioned on these data. An advanced algorithm to select a proper set of nearby views or an adaptive approximation of the optimal number of required cost volumes for a scene could be promising extensions to our method.

\bigbreak\noindent\textbf{Acknowledgement} \\
This research was supported by ams OSRAM.

{\small
\bibliographystyle{ieee_fullname}
\bibliography{egbib}
}

\end{document}


\title{Supplementary Materials for \\ GeoNeRF: Generalizing NeRF with Geometry Priors}

\author{Mohammad Mahdi Johari\\
Idiap Research Institute, EPFL\\
{\tt\small mohammad.johari@idiap.ch}
\and
Yann Lepoittevin\\
ams OSRAM\\
{\tt\small yann.lepoittevin@ams.com}
\and
François Fleuret\\
University of Geneva, EPFL\\
{\tt\small francois.fleuret@unige.ch}
}

\maketitle


\section{Additional Technical Details}

As stated in the main article, we borrow the architecture of our geometry reasoner from CasMVSNet~\cite{gu2020cascade}. We construct $D^{(2)} = 48$ depth planes for the coarsest cost volume, $D^{(1)} = 32$ for the intermediate one, and $D^{(0)} = 8$ for the finest full-resolution cost volume. We use channel size $C = 8$ in group-wise correlation similarity calculations. When training the generalizable model, we create a set of 3$\textrm{--}$5 nearby source views for constructing each cost volume, whereas for fine-tuning and evaluating, we always use a set of 5 nearby views. Also, we scale input images with a factor uniformly sampled from $\{$1.0, 0.75, 0.5$\}$ when we train our generalizable model.

The network architectures of Feature Pyramid Network (FPN), 3D regularizer ($R^{(l)}_{3D}$), and the auto-encoder (AE) are provided in Tables~\ref{table:arch_fpn}, \ref{table:arch_reg3D}, and \ref{table:arch_ae} respectively.

\begin{table}[!b]
    \begin{center}
        \begin{threeparttable}
            \begin{tabular}{lll}
            Input &  Layer & Output  \\
            \hline
            Input & ConvBnReLU(3, 8, 3, 1) & conv0\_0  \\
            conv0\_0 & ConvBnReLU(8, 8, 3, 1) & conv0 \\
            conv0 & ConvBnReLU(8, 16, 5, 2) & conv1\_0 \\
            conv1\_0 & ConvBnReLU(16, 16, 3, 1) & conv1\_1 \\
            conv1\_1 & ConvBnReLU(16, 16, 3, 1) & conv1 \\
            conv1 & ConvBnReLU(16, 32, 5, 2) & conv2\_0 \\
            conv2\_0 & ConvBnReLU(32, 32, 3, 1) & conv2\_1 \\
            conv2\_1 & ConvBnReLU(32, 32, 3, 1) & conv2 \\
            conv2 & Conv(32, 32, 1, 1) & feat2 \\
            conv1 & Conv(16, 32, 1, 1) & f1\_0 \\
            conv0 & Conv(8, 32, 1, 1) & f0\_0 \\
            (feat2, f1\_0) & Upsample\_and\_Add($x$, $y$) & f1\_1 \\
            (f1\_1, f0\_0) & Upsample\_and\_Add($x$, $y$) & f0\_1 \\
            f1\_1 & Conv(32, 16, 3, 1) & feat1 \\
            f0\_1 & Conv(32, 8, 3, 1) & feat0 \\
            \hline
            \end{tabular}
        \end{threeparttable}
    \end{center}
    \vspace{-1.5ex}
    \caption{Network architecture of Feature Pyramid Network (FPN), where \textit{feat2}, \textit{feat1}, and \textit{feat0} are output feature pyramids. Conv($c_{in}$, $c_{out}$, $k$, $s$) stands for a 2D convolution with input channels~$c_{in}$, output channels~$c_{out}$, kernel size of~$k$, and stride of~$s$. ConvBnReLU represents a Conv layer followed by Batch Normalization and ReLU nonlinearity. Upsample\_and\_Add($x$, $y$) adds $y$ to the bilinearly upsampled of $x$.}
    \vspace{1ex}
    \label{table:arch_fpn}
\end{table}

\section{Additional Qualitative Analysis}
\begin{figure*}
    \begin{center}
        \includegraphics[width=0.99\linewidth]{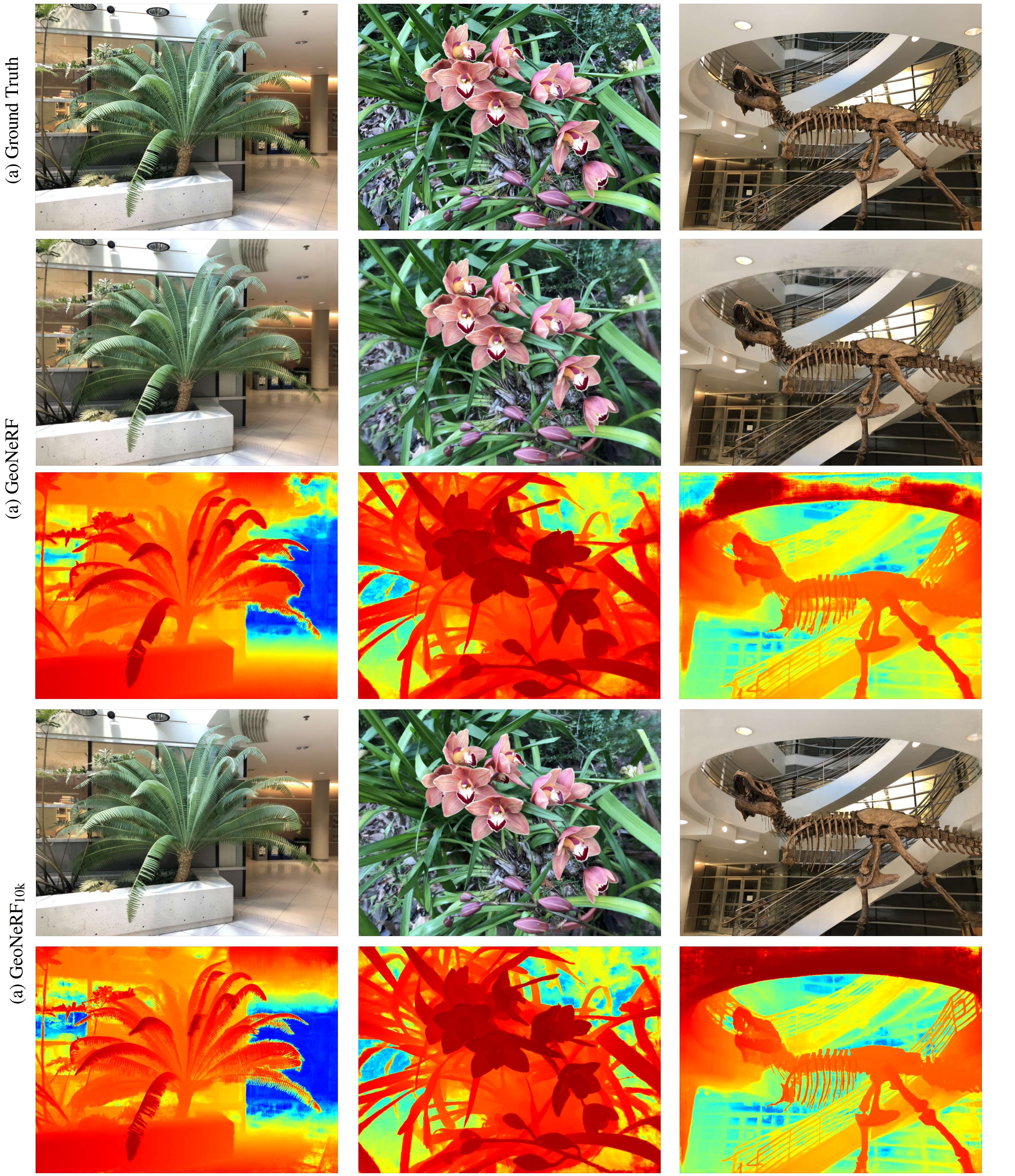}
    \end{center}
    \vspace{-3.25ex}
   \caption{Full-size examples of novel images and their depth map rendered by our generalizable (GeoNeRF) and fine-tuned ($\text{GeoNeRF}_{\text{10k}}$) models. The images are from test scenes of the real forward-facing LLFF dataset~\cite{mildenhall2019llff}.}
   \vspace{-2.5ex}
    \label{fig:supp_llff}
\end{figure*}

\begin{figure*}
    \begin{center}
        \includegraphics[width=0.99\linewidth]{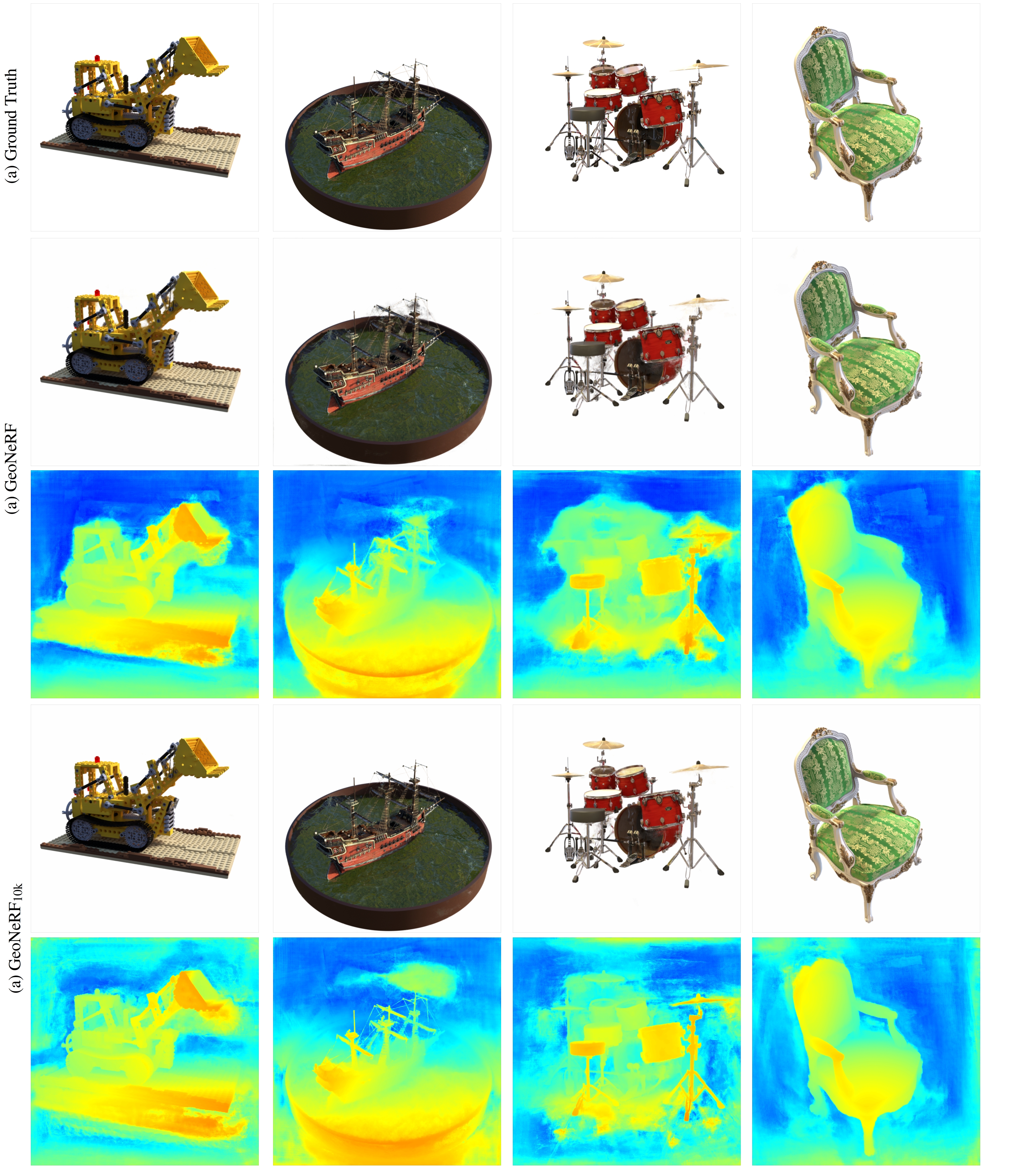}
    \end{center}
    \vspace{-3.25ex}
   \caption{Full-size examples of novel images and their depth map rendered by our generalizable (GeoNeRF) and fine-tuned ($\text{GeoNeRF}_{\text{10k}}$) models. The images are from test scenes of the NeRF realistic synthetic dataset~\cite{mildenhall2020nerf}.}
   \vspace{-2.5ex}
    \label{fig:supp_nerf}
\end{figure*}

Full-size examples of rendered images for novel views by our GeoNeRF model are presented in Figures~\ref{fig:supp_llff} and \ref{fig:supp_nerf}. Figure~\ref{fig:supp_llff} includes samples from the real forward-facing LLFF dataset~\cite{mildenhall2019llff}, and Figure~\ref{fig:supp_nerf} contains samples from the NeRF realistic synthetic dataset\cite{mildenhall2020nerf}. In addition to the rendered images, we also show the rendered depth maps for each novel view. Images indicated by GeoNeRF are rendered by our generalizable model, while images indicated by $\text{GeoNeRF}_{\text{10k}}$ are rendered after fine-tuning our model on each scene for 10k iterations.

\section{Per-Scene Breakdown}
Tables~\ref{table:per_scene_no_ft_llff}, \ref{table:per_scene_ft_llff}, \ref{table:per_scene_no_ft_nerf}, and \ref{table:per_scene_ft_nerf} break down the quantitative results presented in the main paper into per-scene metrics. The results are consistent with the aggregate results in the main paper.  Tables~\ref{table:per_scene_no_ft_llff} and \ref{table:per_scene_ft_llff} include the scenes from the real forward-facing LLFF dataset~\cite{mildenhall2019llff}, and Tables~\ref{table:per_scene_no_ft_nerf} and \ref{table:per_scene_ft_nerf} contain the scenes from NeRF realistic synthetic dataset~\cite{mildenhall2020nerf}. As it was already shown in the main paper, our generalizable GeoNeRF model outperforms all existing generalizable methods on average, and after fine-tuning, it is on par with per-scene optimized vanilla NeRF~\cite{mildenhall2020nerf}.

\begin{table}[!t]
    \begin{center}
        \begin{threeparttable}
            \begin{tabular}{lll}
            Input &  Layer & Output  \\
            \hline
            Input & ConvBnReLU(8, 8, 3, 1) & conv0  \\
            conv0 & ConvBnReLU(8, 16, 3, 2) & conv1 \\
            conv1 & ConvBnReLU(16, 16, 3, 1) & conv2 \\
            conv2 & ConvBnReLU(16, 32, 3, 2) & conv3 \\
            conv3 & ConvBnReLU(32, 32, 3, 1) & conv4 \\
            conv4 & ConvBnReLU(32, 64, 3, 2) & conv5 \\
            conv5 & ConvBnReLU(64, 64, 3, 1) & conv6 \\
            conv6 & TrpsConvBnReLU(64, 32, 3, 2) & x\_0 \\
            (conv4, x\_0) & Add($x$, $y$) & x\_1 \\
            x\_1 & TrpsConvBnReLU(32, 16, 3, 2) & x\_2 \\
            (conv2, x\_2) & Add($x$, $y$) & x\_3 \\
            x\_3 & TrpsConvBnReLU(16, 8, 3, 2) & x\_4 \\
            (conv0, x\_4) & Add($x$, $y$) & x\_5 \\
            x\_5 & ConvBnReLU(8, 8, 3, 1) & prob\_0  \\
            prob\_0 & Conv(8, 1, 3, 1) & prob  \\
            x\_5 & ConvBnReLU(8, 8, 3, 1) & feat  \\
            \hline
            \end{tabular}
        \end{threeparttable}
    \end{center}
    \vspace{-1.5ex}
    \caption{Network architecture of the 3D regularizer ($R^{(l)}_{3D}$), where \textit{feat} is the output 3D feature map~$\boldsymbol{\Phi^{(l)}}$ and \textit{prob} is the output probability which is used to regress the depth map~$\boldsymbol{\hat{D}^{(l)}}$. Conv($c_{in}$, $c_{out}$, $k$, $s$) stands for a 3D convolution with input channels~$c_{in}$, output channels~$c_{out}$, kernel size of~$k$, and stride of~$s$. ConvBnReLU represents a Conv layer followed by Batch Normalization and ReLU nonlinearity, and TrpsConv stands for transposed 3D convolution. Add($x$, $y$) simply adds $y$ to $x$.}
    \vspace{1ex}
    \label{table:arch_reg3D}
\end{table}

\begin{table}[!t]
    \begin{center}
        \begin{threeparttable}
            \begin{tabular}{lll}
            Input &  Layer & Output  \\
            \hline
            Input & ConvLnELU(32, 64, 3, 1) & conv1\_0  \\
            conv1\_0 & MaxPool & conv1 \\
            conv1 & ConvLnELU(64, 128, 3, 1) & conv2\_0 \\
            conv2\_0 & MaxPool & conv2 \\
            conv2 & ConvLnELU(128, 128, 3, 1) & conv3\_0 \\
            conv3\_0 & MaxPool & conv3 \\
            conv3 & TrpsConvLnELU(128, 128, 4, 2) & x\_0 \\
            $[$ conv2 ; x\_0 $]$ & TrpsConvLnELU(256, 64, 4, 2) & x\_1 \\
            $[$ conv1 ; x\_1 $]$ & TrpsConvLnELU(128, 32, 4, 2) & x\_2 \\
            $[$ Input ; x\_2 $]$ & ConvLnELU(64, 32, 3, 1) & Output \\
            \hline
            \end{tabular}
        \end{threeparttable}
    \end{center}
    \vspace{-1.5ex}
    \caption{Network architecture of the auto-encoder network (AE). Conv($c_{in}$, $c_{out}$, $k$, $s$) stands for a 1D convolution with input channels~$c_{in}$, output channels~$c_{out}$, kernel size of~$k$, and stride of~$s$. ConvLnELU represents a Conv layer followed by Layer Normalization and ELU nonlinearity, and TrpsConv stands for transposed 1D convolution. MaxPool is a 1D max pooling layer with a stride of~2, and $[\cdot$ ; $\cdot]$ denotes concatenation.}
    \vspace{1ex}
    \label{table:arch_ae}
\end{table}

\clearpage

\begin{table*}[!t]
    \begin{center}
        \begin{threeparttable}
            \begin{tabular}{l|cccccccccc}
            & \multicolumn{8}{c}{PSNR$\uparrow$} \\
            &  Fern & Flower & Fortress & Horns & Leaves & Orchids & Room & T-Rex \\
            \hline
            pixelNeRF~\cite{yu2021pixelnerf} & 12.40 & 10.00 & 14.07 & 11.07 & \phantom{0}9.85 & \phantom{0}9.62 & 11.75 & 10.55 \\
            IBRNet~\cite{wang2021ibrnet} & 23.84 & 26.67 & 30.00 & 26.48 & 20.19 & 19.34 & \textbf{29.94} & \textbf{24.57} \\
            MVSNeRF~\cite{chen2021mvsnerf} & 21.15 & 24.74 & 26.03 & 23.57 & 17.51 & 17.85 & 26.95 & 23.20 \\
            GeoNeRF & \textbf{24.61} & \textbf{28.12} & \textbf{30.49} & \textbf{26.96} & \textbf{20.58} & \textbf{20.24} & 28.74 & 23.75 \\
            \hline
            \end{tabular}
        \end{threeparttable}
    \end{center}
    \vspace{-2.5ex}
\end{table*}

\begin{table*}[!t]
    \begin{center}
        \begin{threeparttable}
            \begin{tabular}{l|cccccccccc}
            & \multicolumn{8}{c}{SSIM$\uparrow$} \\
            &  Fern & Flower & Fortress & Horns & Leaves & Orchids & Room & T-Rex \\
            \hline
            pixelNeRF~\cite{yu2021pixelnerf} & 0.531 & 0.433 & 0.674 & 0.516 & 0.268 & 0.317 & 0.691 & 0.458 \\
            IBRNet~\cite{wang2021ibrnet} & 0.772 & 0.856 & 0.883 & 0.869 & 0.719 & 0.633 & 0.946 & 0.861 \\
            MVSNeRF~\cite{chen2021mvsnerf} & 0.638 & \textbf{0.888} & 0.872 & 0.868 & 0.667 & 0.657 & \textbf{0.951} & 0.868 \\
            GeoNeRF & \textbf{0.811} & 0.885 & \textbf{0.898} & \textbf{0.901} & \textbf{0.741} & \textbf{0.666} & 0.935 & \textbf{0.877} \\
            \hline
            \end{tabular}
        \end{threeparttable}
    \end{center}
    \vspace{-2.5ex}
\end{table*}

\begin{table*}[!t]
    \begin{center}
        \begin{threeparttable}
            \begin{tabular}{l|cccccccccc}
            & \multicolumn{8}{c}{LPIPS$\downarrow$} \\
            &  Fern & Flower & Fortress & Horns & Leaves & Orchids & Room & T-Rex \\
            \hline
            pixelNeRF~\cite{yu2021pixelnerf} & 0.650 & 0.708 & 0.608 & 0.705 & 0.695 & 0.721 & 0.611 & 0.667 \\
            IBRNet~\cite{wang2021ibrnet} & 0.246 & 0.164 & 0.153 & 0.177 & 0.230 & 0.287 & 0.153 & 0.230 \\
            MVSNeRF~\cite{chen2021mvsnerf} & 0.238 & 0.196 & 0.208 & 0.237 & 0.313 & 0.274 & 0.172 & \textbf{0.184} \\
            GeoNeRF & \textbf{0.202} & \textbf{0.133} & \textbf{0.123} & \textbf{0.140} & \textbf{0.222} & \textbf{0.256} & \textbf{0.150} & 0.212 \\
            \hline
            \end{tabular}
        \end{threeparttable}
    \end{center}
    \vspace{-1.5ex}
    \caption{Per-scene Quantitative comparison of our proposed GeoNeRF with existing generalizable NeRF models on real forward-facing LLFF dataset~\cite{mildenhall2019llff} in terms of PSNR (higher is better), SSIM~\cite{wang2004image} (higher is better), and LPIPS~\cite{zhang2018unreasonable} (lower is better) metrics.}
    \vspace{1ex}
    \label{table:per_scene_no_ft_llff}
\end{table*}

\begin{table*}[!t]
    \begin{center}
        \begin{threeparttable}
            \begin{tabular}{l|cccccccccc}
            & \multicolumn{8}{c}{PSNR$\uparrow$} \\
            &  Fern & Flower & Fortress & Horns & Leaves & Orchids & Room & T-Rex \\
            \hline
            NeRF~\cite{mildenhall2020nerf} & 25.17 & 27.40 & \textbf{31.16} & 27.45 & 20.92 & 20.36 & \textbf{32.70} & \textbf{26.80} \\
            $\text{GeoNeRF}_{\text{10k}}$ & \textbf{25.24} & 28.57 & 30.75 & \textbf{28.12} & \textbf{21.40} & 20.39 & 31.51 & 26.63 \\
            $\text{GeoNeRF}_{\text{1k}}$ & 25.08 & \textbf{28.74} & 30.83 & 27.66 & 21.16 & \textbf{20.41} & 30.52 & 26.07\\            
            \hline
            \end{tabular}
        \end{threeparttable}
    \end{center}
    \vspace{-2.5ex}
\end{table*}

\begin{table*}[!t]
    \begin{center}
        \begin{threeparttable}
            \begin{tabular}{l|cccccccccc}
            & \multicolumn{8}{c}{SSIM$\uparrow$} \\
            &  Fern & Flower & Fortress & Horns & Leaves & Orchids & Room & T-Rex \\
            \hline
            NeRF~\cite{mildenhall2020nerf} & 0.792 & 0.827 & 0.881 & 0.828 & 0.690 & 0.641 & 0.948 & 0.880 \\
            $\text{GeoNeRF}_{\text{10k}}$ & \textbf{0.829} & 0.890 & 0.900 & \textbf{0.912} & \textbf{0.781} & \textbf{0.674} & \textbf{0.956} & \textbf{0.910} \\
            $\text{GeoNeRF}_{\text{1k}}$ & 0.824 & \textbf{0.892} & \textbf{0.905} & 0.908 & 0.769 & 0.673 & 0.946 & 0.901 \\            
            \hline
            \end{tabular}
        \end{threeparttable}
    \end{center}
    \vspace{-2.5ex}
\end{table*}

\begin{table*}[!t]
    \begin{center}
        \begin{threeparttable}
            \begin{tabular}{l|cccccccccc}
            & \multicolumn{8}{c}{LPIPS$\downarrow$} \\
            &  Fern & Flower & Fortress & Horns & Leaves & Orchids & Room & T-Rex \\
            \hline
            NeRF~\cite{mildenhall2020nerf} & 0.280 & 0.219 & 0.171 & 0.268 & 0.316 & 0.321 & 0.178 & 0.249 \\
            $\text{GeoNeRF}_{\text{10k}}$ & \textbf{0.185} & 0.120 & 0.125 & \textbf{0.126} & \textbf{0.183} & \textbf{0.247} & \textbf{0.126} & \textbf{0.181} \\
            $\text{GeoNeRF}_{\text{1k}}$ & 0.189 & \textbf{0.114} & \textbf{0.117} & 0.130 & 0.198 & 0.248 & 0.135 & 0.188 \\            
            \hline
            \end{tabular}
        \end{threeparttable}
    \end{center}
    \vspace{-1.5ex}
    \caption{Per-scene Quantitative comparison of our fine-tuned GeoNeRF with per-scene optimized vanilla NeRF~\cite{mildenhall2020nerf} on real forward-facing LLFF dataset~\cite{mildenhall2019llff} in terms of PSNR (higher is better), SSIM~\cite{wang2004image} (higher is better), and LPIPS~\cite{zhang2018unreasonable} (lower is better) metrics. Our model is fine-tuned on each scene for 10k iterations ($\text{GeoNeRF}_{\text{10k}}$) and 1k iterations ($\text{GeoNeRF}_{\text{1k}}$), and NeRF~\cite{mildenhall2020nerf} is optimized for 200k iterations.}
    \label{table:per_scene_ft_llff}
\end{table*}

\clearpage

\begin{table*}[!t]
    \begin{center}
        \begin{threeparttable}
            \begin{tabular}{l|cccccccccc}
            & \multicolumn{8}{c}{PSNR$\uparrow$} \\
            &  Chair & Drums & Ficus & Hotdog & Lego & Materials & Mic & Ship \\
            \hline
            pixelNeRF~\cite{yu2021pixelnerf} & \phantom{0}7.18 & \phantom{0}8.15 & \phantom{0}6.61 & \phantom{0}6.80 & \phantom{0}7.74 & \phantom{0}7.61 & \phantom{0}7.71 & \phantom{0}7.30 \\
            IBRNet~\cite{wang2021ibrnet} & 28.54 & 21.22 & 24.23 & 31.72 & 24.59 & 22.20 & 27.97 & 23.64 \\
            MVSNeRF~\cite{chen2021mvsnerf} & 23.35 & 20.71 & 21.98 & 28.44 & 23.18 & 20.05 & 22.62 & 23.35 \\
            GeoNeRF & \textbf{31.84} & \textbf{24.00} & \textbf{25.28} & \textbf{34.33} & \textbf{28.80} & \textbf{26.16} & \textbf{31.15} & \textbf{25.08} \\
            \hline
            \end{tabular}
        \end{threeparttable}
    \end{center}
    \vspace{-2.5ex}
\end{table*}

\begin{table*}[!t]
    \begin{center}
        \begin{threeparttable}
            \begin{tabular}{l|cccccccccc}
            & \multicolumn{8}{c}{SSIM$\uparrow$} \\
            &  Chair & Drums & Ficus & Hotdog & Lego & Materials & Mic & Ship \\
            \hline
            pixelNeRF~\cite{yu2021pixelnerf} & 0.624 & 0.670 & 0.669 & 0.669 & 0.671 & 0.644 & 0.729 & 0.584 \\
            IBRNet~\cite{wang2021ibrnet} & 0.948 & 0.896 & 0.915 & 0.952 & 0.918 & 0.905 & 0.962 & 0.834 \\
            MVSNeRF~\cite{chen2021mvsnerf} & 0.876 & 0.886 & 0.898 & 0.962 & 0.902 & 0.893 & 0.923 & \textbf{0.886} \\
            GeoNeRF & \textbf{0.973} & \textbf{0.921} & \textbf{0.931} & \textbf{0.975} & \textbf{0.956} & \textbf{0.926} & \textbf{0.978} & 0.844 \\
            \hline
            \end{tabular}
        \end{threeparttable}
    \end{center}
    \vspace{-2.5ex}
\end{table*}

\begin{table*}[!t]
    \begin{center}
        \begin{threeparttable}
            \begin{tabular}{l|cccccccccc}
            & \multicolumn{8}{c}{LPIPS$\downarrow$} \\
            &  Chair & Drums & Ficus & Hotdog & Lego & Materials & Mic & Ship \\
            \hline
            pixelNeRF~\cite{yu2021pixelnerf} & 0.386 & 0.421 & 0.335 & 0.433 & 0.427 & 0.432 & 0.329 & 0.526 \\
            IBRNet~\cite{wang2021ibrnet} & 0.066 & \textbf{0.091} & 0.097 & 0.067 & 0.095 & \textbf{0.115} & 0.051 & 0.219 \\
            MVSNeRF~\cite{chen2021mvsnerf} & 0.282 & 0.187 & 0.211 & 0.173 & 0.204 & 0.216 & 0.177 & 0.244 \\
            GeoNeRF & \textbf{0.040} & 0.098 & \textbf{0.092} & \textbf{0.056} & \textbf{0.059} & 0.116 & \textbf{0.037} & \textbf{0.200} \\
            \hline
            \end{tabular}
        \end{threeparttable}
    \end{center}
    \vspace{-1.5ex}
    \caption{Per-scene Quantitative comparison of our proposed GeoNeRF with existing generalizable NeRF models on NeRF realistic synthetic dataset~\cite{mildenhall2020nerf} in terms of PSNR (higher is better), SSIM~\cite{wang2004image} (higher is better), and LPIPS~\cite{zhang2018unreasonable} (lower is better) metrics.}
    \vspace{1ex}
    \label{table:per_scene_no_ft_nerf}
\end{table*}

\begin{table*}[!t]
    \begin{center}
        \begin{threeparttable}
            \begin{tabular}{l|cccccccccc}
            & \multicolumn{8}{c}{PSNR$\uparrow$} \\
            &  Chair & Drums & Ficus & Hotdog & Lego & Materials & Mic & Ship \\
            \hline
            NeRF~\cite{mildenhall2020nerf} & 33.00 & 25.01 & \textbf{30.13} & 36.18 & \textbf{32.54} & \textbf{29.62} & 32.91 & 28.65 \\
            $\text{GeoNeRF}_{\text{10k}}$ & \textbf{33.54} & \textbf{25.13} & 27.79 & \textbf{36.26} & 30.32 & 28.19 & \textbf{33.41} & \textbf{28.76} \\
            $\text{GeoNeRF}_{\text{1k}}$ & 32.76 & 24.74 & 27.06 & 35.71 & 29.79 & 27.69 & 32.83 & 28.11 \\            
            \hline
            \end{tabular}
        \end{threeparttable}
    \end{center}
    \vspace{-2.5ex}
\end{table*}

\begin{table*}[!t]
    \begin{center}
        \begin{threeparttable}
            \begin{tabular}{l|cccccccccc}
            & \multicolumn{8}{c}{SSIM$\uparrow$} \\
            &  Chair & Drums & Ficus & Hotdog & Lego & Materials & Mic & Ship \\
            \hline
            NeRF~\cite{mildenhall2020nerf} & 0.967 & 0.925 & \textbf{0.964} & 0.974 & 0.961 & 0.949 & 0.980 & 0.856 \\
            $\text{GeoNeRF}_{\text{10k}}$ & \textbf{0.980} & \textbf{0.935} & 0.955 & \textbf{0.983} & \textbf{0.965} & \textbf{0.953} & \textbf{0.987} & \textbf{0.890} \\
            $\text{GeoNeRF}_{\text{1k}}$ & 0.977 & 0.930 & 0.948 & 0.982 & 0.961 & 0.948 & 0.985 & 0.883 \\            
            \hline
            \end{tabular}
        \end{threeparttable}
    \end{center}
    \vspace{-2.5ex}
\end{table*}

\begin{table*}[!t]
    \begin{center}
        \begin{threeparttable}
            \begin{tabular}{l|cccccccccc}
            & \multicolumn{8}{c}{LPIPS$\downarrow$} \\
            &  Chair & Drums & Ficus & Hotdog & Lego & Materials & Mic & Ship \\
            \hline
            NeRF~\cite{mildenhall2020nerf} & 0.046 & 0.091 & \textbf{0.044} & 0.121 & 0.050 & 0.063 & 0.028 & 0.206 \\
            $\text{GeoNeRF}_{\text{10k}}$ & \textbf{0.024} & \textbf{0.073} & 0.061 & \textbf{0.032} & \textbf{0.041} & \textbf{0.058} & \textbf{0.016} & \textbf{0.137} \\
            $\text{GeoNeRF}_{\text{1k}}$ & 0.030 & 0.081 & 0.069 & 0.034 & 0.046 & 0.069 & 0.020 & 0.145 \\            
            \hline
            \end{tabular}
        \end{threeparttable}
    \end{center}
    \vspace{-1.5ex}
    \caption{Per-scene Quantitative comparison of our fine-tuned GeoNeRF with per-scene optimized vanilla NeRF~\cite{mildenhall2020nerf} on NeRF realistic synthetic dataset~\cite{mildenhall2020nerf} in terms of PSNR (higher is better), SSIM~\cite{wang2004image} (higher is better), and LPIPS~\cite{zhang2018unreasonable} (lower is better) metrics. Our model is fine-tuned on each scene for 10k iterations ($\text{GeoNeRF}_{\text{10k}}$) and 1k iterations ($\text{GeoNeRF}_{\text{1k}}$), and NeRF~\cite{mildenhall2020nerf} is optimized for 500k iterations.}
    \label{table:per_scene_ft_nerf}
\end{table*}

\section{Ablation Study} \label{sec:ablation}
An ablation study of our generalizable model on the NeRF synthetic dataset~\cite{mildenhall2020nerf} and the real forward-facing dataset~\cite{mildenhall2019llff} is presented in Table~\ref{table:ablation}, contrasting the effectiveness of individual components of our proposed model. We evaluated GeoNeRF in the cases where (a) no self-supervision loss is used, (b) no positional encoding is employed, (c) points on a ray are merely sampled uniformly, (d) occluded views are not excluded, (e) attention mechanism is removed from the renderer, (f) view-independent tokens are not regularized with the AE network before predicting volume densities, and (g) only a single cost volume is constructed per-view instead of cascaded multi-level cost volumes.

\begin{figure*}
	\begin{center}
		\includegraphics[width=0.99\linewidth]{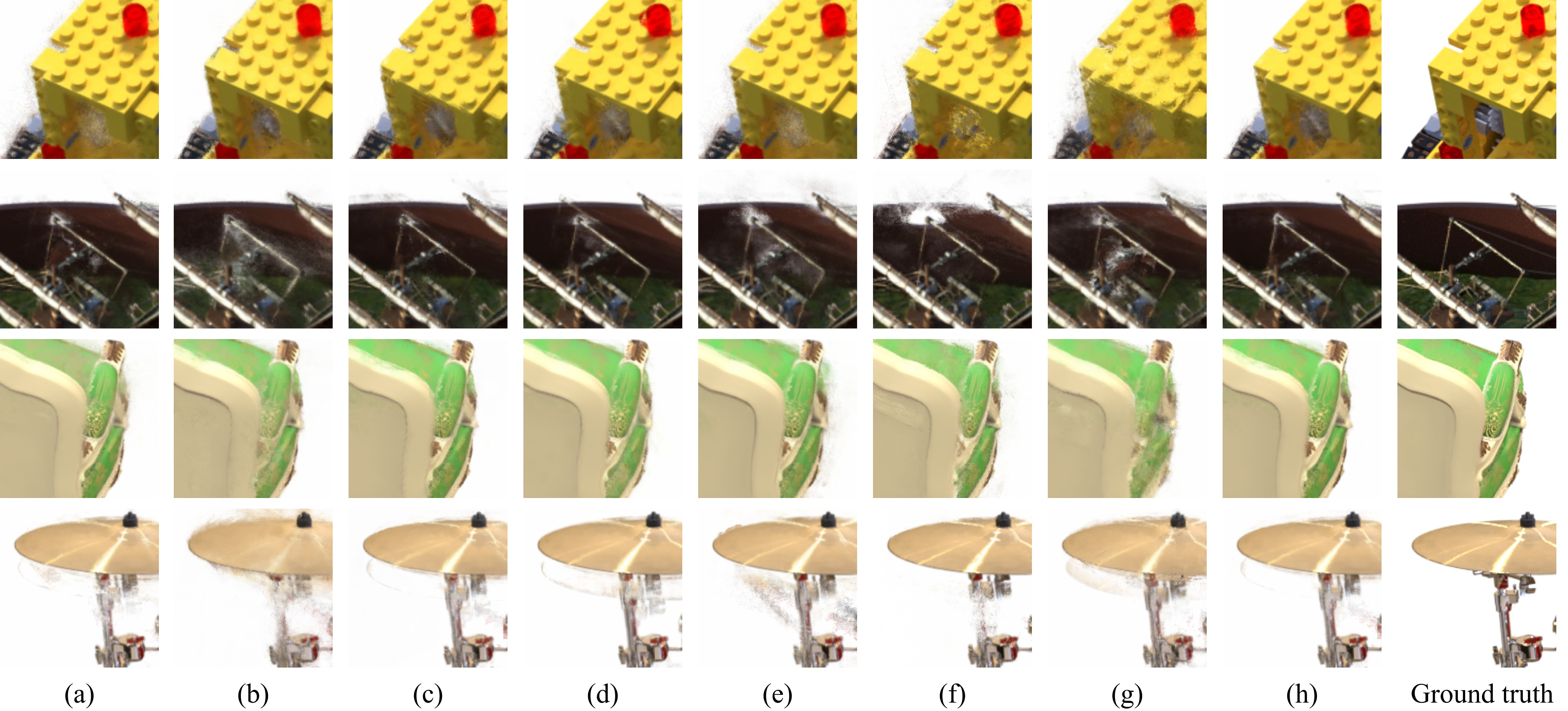}
	\end{center}
	\vspace{-2ex}
	\caption{Qualitative ablation study of the key components of GeoNeRF. The examples are selected from challenging views of the NeRF synthetic dataset~\cite{mildenhall2020nerf}. Columns correspond to the experiments in Table~\ref{table:ablation}.}
	\vspace{-2ex}
	\label{fig:ablation}
\end{figure*}

Figure~\ref{fig:ablation} contains examples from the NeRF synthetic dataset~\cite{mildenhall2020nerf} for qualitative analysis corresponding to the experiments in Table~\ref{table:ablation}. The examples focus on challenging views of the scenes in order to contrast the behavior of the models properly.

\begin{table}[!t]
	\begin{center}
		\begin{tabular}{l|ccc|ccc|c}
			\hline
			\multirow{2}{*}{Experiment} & \multicolumn{3}{c|}{Realistic Synthetic NeRF~\cite{mildenhall2020nerf}} &  \multicolumn{3}{c|}{Real Forward Facing LLFF~\cite{mildenhall2019llff}} & \multirow{2}{*}{Examples} \\
			& PSNR$\uparrow$ &  SSIM$\uparrow$ & LPIPS$\downarrow$ & PSNR$\uparrow$ &  SSIM$\uparrow$ & LPIPS$\downarrow$ \\
			\hline
			a. Without self-supervision & 28.10 & 0.935 & 0.098  & 25.37 & 0.836 & 0.184 & Figure~\ref{fig:ablation}.a \\
			b. Without positional encoding & 27.19 & 0.927 & 0.116 & 25.02 & 0.836 & 0.189 & Figure~\ref{fig:ablation}.b \\
			c. Uniform sampling along a ray & 28.04 & 0.934 & 0.089  & 25.31 & 0.835 & 0.184 & Figure~\ref{fig:ablation}.c \\
			d. Without occlusion masks & 27.92 & 0.932 & 0.097  & 25.22 & 0.834 & 0.185 & Figure~\ref{fig:ablation}.d \\
			e. Without attention mechanism & 27.69 & 0.929 & 0.135 & 24.95 & 0.828 & 0.194 & Figure~\ref{fig:ablation}.e \\
			f. Without the AE network & 23.53 & 0.884 & 0.182 & 24.92 & 0.821 & 0.199 & Figure~\ref{fig:ablation}.f \\			
			g. Single cost volume & 26.60 & 0.915 & 0.132 & 24.60 & 0.814 & 0.211 & Figure~\ref{fig:ablation}.g \\
			h. Full GeoNeRF & \textbf{28.33} & \textbf{0.938} & \textbf{0.087} & \textbf{25.44} & \textbf{0.839} & \textbf{0.180} & Figure~\ref{fig:ablation}.h \\
			\hline
		\end{tabular}
	\end{center}
	\vspace{-3ex}
	\caption{Ablation study of the key components of GeoNeRF. The evaluation is performed on the NeRF synthetic~\cite{mildenhall2020nerf} and the real forward-facing LLFF~\cite{mildenhall2019llff} test scenes. See Section~\ref{sec:ablation} for the details of these experiments, and see Figure~\ref{fig:ablation} for qualitative analysis.}
	\label{table:ablation}
\end{table}

\section{Limitations} \label{sec:limitations}
\begin{figure}[!t]
	\begin{center}
		\includegraphics[width=0.65\linewidth]{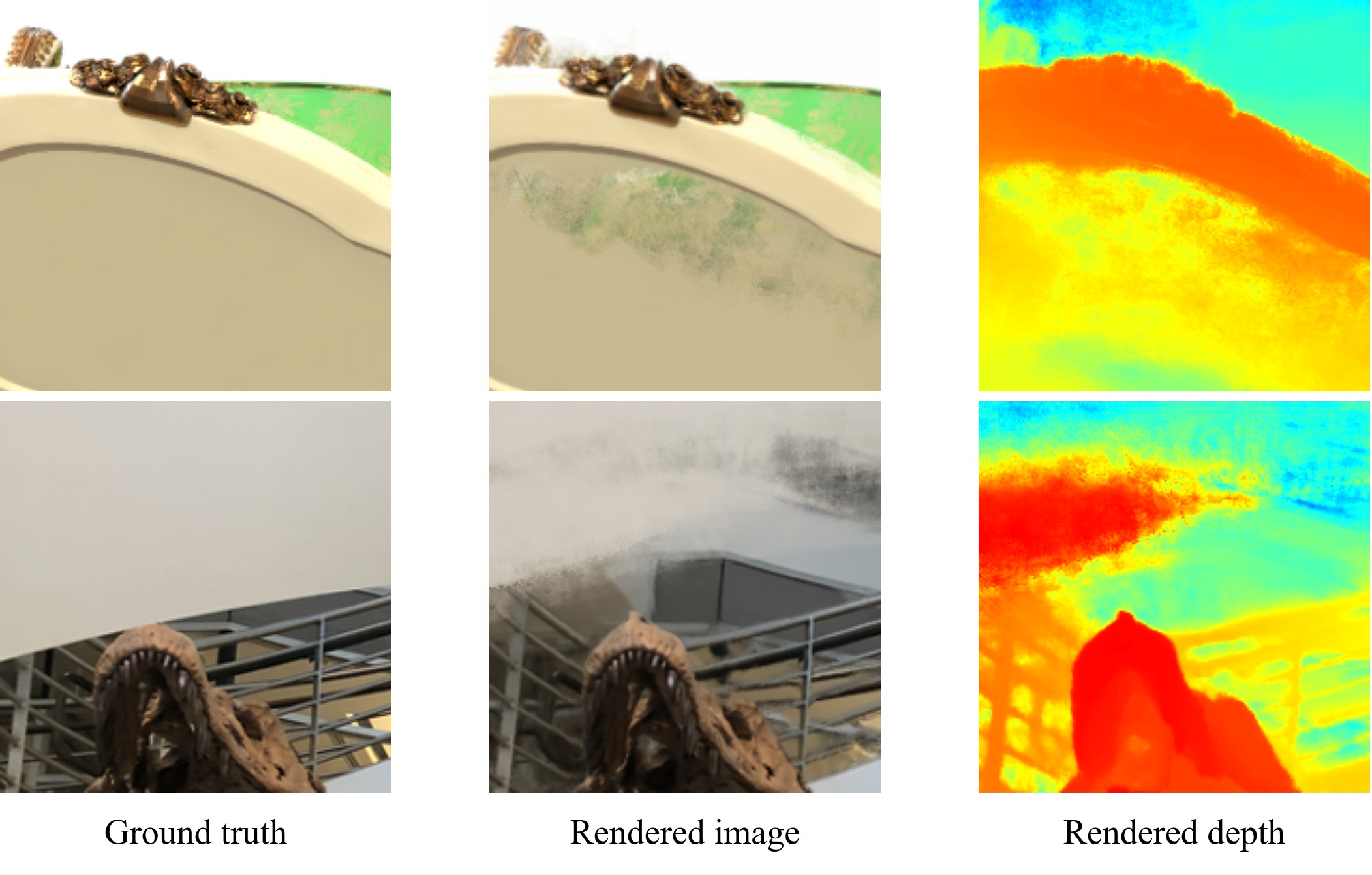}
	\end{center}
	\vspace{-2ex}
	\caption{Failure examples in our method where stereo reconstruction fails in the geometry reasoner for textureless areas.}
	\label{fig:failure}
\end{figure}

Our model with the experimental settings in the main article can be trained and evaluated on a single GPU with 16 GB of memory. Failure cases in our model could occur when the stereo reconstruction fails in the geometry reasoner, and the renderer is misled by incorrect geometry priors. Since the architecture of the geometry reasoner is inspired by multi-view stereo models, it is prone to failure in textureless areas similarly. Such failure examples are shown in Fig.~\ref{fig:failure}.

\clearpage

{\small
\bibliographystyle{ieee_fullname}
\bibliography{egbib}
}